# Promoting the Responsible Development of Speech Datasets for Mental Health and Neurological Disorders Research

**Eleonora Mancini**                                      E.MANCINI@UNIBO.IT
*Department of Computer Science and Engineering,*
*University of Bologna, Bologna, Italy*

**Ana Tanevska** ✉                                      ANA.TANEVSKA@IT.UU.SE
*Department of Information Technology,*
*Uppsala University, Uppsala, Sweden*

**Andrea Galassi** ✉                                      A.GALASSI@UNIBO.IT
*Department of Computer Science and Engineering,*
*University of Bologna, Bologna, Italy*

**Alessio Galatolo**                                      ALESSIO.GALATOLO@IT.UU.SE
*Department of Information Technology,*
*Uppsala University, Uppsala, Sweden*

**Federico Ruggeri**                                      FEDERICO.RUGGERI6@UNIBO.IT
**Paolo Torroni**                                      P.TORRONI@UNIBO.IT
*Department of Computer Science and Engineering,*
*University of Bologna, Bologna, Italy*

## Abstract

Current research in machine learning and artificial intelligence is largely centered on modeling and performance evaluation, less so on data collection. However, recent research demonstrated that limitations and biases in data may negatively impact trustworthiness and reliability. These aspects are particularly impactful on sensitive domains such as mental health and neurological disorders, where speech data are used to develop AI applications for patients and healthcare providers. In this paper, we chart the landscape of available speech datasets for this domain, to highlight possible pitfalls and opportunities for improvement and promote fairness and diversity. We present a comprehensive list of desiderata for building speech datasets for mental health and neurological disorders and distill it into an actionable checklist focused on ethical concerns to foster more responsible research.

## 1. Introduction

Conventional AI-system development follows a model-centric lifecycle, whereby researchers or practitioners iterate the model to improve performance, while the benchmark dataset remains mostly unchanged (Zha et al., 2023). A recent trend recognizes that data collection and curation is a pivotal yet frequently overlooked step in machine learning pipelines and advocates a fundamental shift from model design to data quality and reliability. Issues originating from problematic data collection practices have been demonstrated across various domains (Paullada et al., 2021; Santy et al., 2023; Peng et al., 2021). Non-representative or inaccurate training data can profoundly impact model performance, propagating biases into subsequent development stages, ultimately leading to negative societal impacts (Ntoutsi et al., 2020; Angwin et al., 2022; Papakyriakopoulos et al., 2023). Furthermore, neglecting to follow adequate policies and protocols for collecting the data, coupled with these data biases, can end up compromising the system's overall reliability and trustworthiness (Varona & Suárez, 2022; Liang et al., 2022).

This issue is especially relevant in domains targeting *mental health and neurological disorders* (MHND),[1] where data may be used to develop AI systems that identify high-risk populations for early intervention, detect and predict disorders in decision support systems, and trigger targeted actions, potentially improving access to mental health care (Ettman & Galea, 2023). For instance, digitally-delivered targeted interventions could mitigate the burden of mental illness in hard-to-reach populations. However, creating datasets for MHND applications may be potentially harmful for all the stakeholders. For instance, individuals could volunteer sensitive content when speaking spontaneously during data collection. Such risks must be accounted for and dealt with systematically, for example by obtaining proper participant consent, ensuring data security and confidentiality, and mitigating biases (Fadda et al., 2022; Martinez-Martin et al., 2021; Ettman & Galea, 2023). Moreover, to enable broader accessibility of the datasets, a delicate balance must be maintained between privacy protection and transparent data utilization (Papakyriakopoulos et al., 2023). Recognizing the growing concern surrounding these sensitive topics, previous research has addressed ethical issues at different stages of the development process (Andrews et al., 2023; Gooding & Kariotis, 2021; Terra et al., 2023). However, there is a noticeable gap in the literature concerning speech datasets for MHND.

Building on recent analyses of ethical concerns in MHND and known best practices for data collection, we define a set of desired features and attributes to guide future speech data gathering efforts in MHND applications. We focus on ethical concerns arising in data collection and reflected in data utilization. Our list of desiderata could be used for two purposes. It could be used as an analytical tool by researchers committed to a more aware use of existing resources/results and as a guide for future research. As to the latter point, with reference to the process described in Zhao et al. (2024a), our work aims to support the design, implementation, and maintenance phases of dataset construction. To this end, we translate our list of desiderata into an actionable checklist. Finally, we use the checklist to critically review the literature on MHND speech datasets, identifying pitfalls and highlighting areas for improvement. Our analysis is focused on MHND but it largely applies to other speech tasks as well.

To summarize, our contributions are:

- a comprehensive list of desiderata for building speech datasets concerning MHND, marking the first exploration in this domain;

- a distilled checklist that encodes our desiderata and can serve as a guideline for future research;

- a survey of 36 papers on this domain, where we leverage our checklist to discuss current practices and pinpoint areas for improvement.

---

1. In the clinical literature, a clear distinction between symptoms and disorders is often delineated, yet the majority of datasets in the domain of MHND have primarily focused on detecting disorders, albeit with some contributions towards symptom detection, such as aphasia (Low et al., 2020).

The paper is structured as follows. In Section 2, we motivate our work by discussing a case study. In Section 3, we examine related work. In Section 4, we present and motivate our desiderata. In Section 5, we describe our methodology and checklist. We survey existing literature and discuss our findings in Section 6. Section 7 concludes.

All the material used for our work is made publicly available.[2]

## 2. A Case Study on DAIC-WoZ Depression Research

To motivate our approach, we start with a concrete case study. We select DAIC-WoZ (Gratch et al., 2014) as one of the oldest and arguably most popular datasets regarding depression, which had a pivotal role in the growth of the field and related research.

DAIC-WoZ is a subset of the larger Distress Analysis Interview Corpus (DAIC) (Gratch et al., 2014) where a person is participating in a semi-structured interview to assess their risk for post-traumatic stress disorder (PTSD) and major depression disorder (MDD). The datasets consist of the data from the interviews, such as the speech data and the recordings' transcriptions, and is annotated with the Patient Health Questionnaire (PHQ) for depression (Kroenke & Spitzer, 2002). The dataset has been vastly used for the training of systems aimed at automatic depression detection (Bailey & Plumbley, 2021; Ravi et al., 2022; Yang et al., 2023; Ehghaghi et al., 2022; Jo & Kwak, 2022). In these works, authors develop methods to automatically recognize depression (primarily as a binary classification task) from the audio of the interviews.

DAIC-WoZ presents an imbalance both in terms of class distribution[3] (65% negative, 35% positive) as well as gender/sex distribution (54% females, 46% males). Bias and lack of diversity in a dataset may propagate to the models using them, often compromising their generalization abilities. Bailey and Plumbley (2021) analyze the effect of gender/sex bias on model performance and show how some features like mel-spectrogram, which are commonly used in several works, are not robust to it. Although it may be possible to partially address these issues through ad-hoc methods (Bailey & Plumbley, 2021; Mancini et al., 2024), a more effective approach would be to adopt a rigorous and well-balanced data collection process from the start.

Another issue concerns the meaning of annotations. DAIC-WoZ includes two labels: the PHQ-8 score (Kroenke & Spitzer, 2002) and a corresponding binary label for depression. In the paper presenting DAIC-WoZ, Gratch et al. (2014) do not specify the criterion used to map the PHQ-8 score into the binary label.[4] While most research involving this dataset has followed Kroenke et al. (2009) and used the label "depressed" for instances with scores > 9, Yang et al. (2023) categorize as "depressed" instances with a score > 10, obtaining results that are not directly comparable with previous literature. Clarity in the definition and motivation of the labels used in the dataset would favor future researchers in coherently structuring the task and may prevent inconsistencies in the experimental results. This would especially favor researchers who do not have a medical background.

---

A final issue worth mentioning is maintainability. Often datasets have known issues with some instances. DAIC-WoZ is no exception. For example, Bailey and Plumbley (2021) identify labeling errors, out-of-sync segments, and missing transcriptions.[5] Errors are difficult to avoid and may only become evident after the dataset has been made public. However, a solid distribution infrastructure supporting maintainability may help address this problem effectively. To the best of our knowledge, there is no method to report these issues or update the dataset with corrections, which inevitably puts unaware researchers and general users of the dataset in a disadvantaged position.

These limitations of DAIC-WoZ, which have emerged over many years of use of the dataset, outline examples of problems that may also be present in other datasets. Therefore, the definition of best practices and desiderata for speech datasets in the MHND domain could induce a principled approach to creating new resources and a more aware use of existing ones.

## 3. Related Work

To the best of our knowledge, there are no desiderata covering the creation and collection of speech datasets in MHND research. However, a rich literature offers analyses of ethical concerns and best practices for dataset construction in related domains.

### 3.1 Ethical Concerns in MHND Research

Working at the intersection of social media and mental health, Fadda et al. (2022) cover the potential ethical concerns when collecting, storing, and analyzing geo-referenced tweets. They identify four issues and suggested methods to address them. They are: (1) *users' privacy expectations and informed consent* - stressing the importance of informed consent when collecting data as well as ensuring that participants understand the implications of sharing data in a public online domain; (2) *data security* - focusing on technological solutions like encryption; (3) *data confidentiality* - related to (1), this point also focuses on the importance of informing participants how their data is stored and transferred, as well as the potential risk of breaching the data confidentiality; and (4) *social responsibility* - advocating for special consideration and respect towards social groups and structures that may be disrupted by the proposed research (e.g. stigmatizing users by identifying them as diagnosed with a mental condition).

Martinez-Martin et al. (2021) extend the scope of the study from social media to a broaden set of phenotyping tools for mental health applications, including biometric and personal data from smartphones and wearables. In their analysis of ethical concerns in this field, they relied a panel of experts in digital phenotyping, data science, mental health, law, and ethics. The following concerns emerged: (1) *privacy and data protection*; (2) *transparency* of the research processes, risks, limitations, and results; (3) informed *consent* of everyone whose personal data is collected; (4) *oversight and accountability* by way of external ethical reviews; (5) *fairness and bias*; and (6) *validation of tools* by establishing the evidence of validity for the use of said tools.

---

5. https://github.com/adbailey1/daic_woz_process

A more broad analysis in the ethics and law in technological research in mental health care is provided by Gooding and Kariotis (2021). They surveyed empirical scholarly literature on the application of algorithmic and data-driven technologies (including social media, smartphones, sensing technology, and chatbots) in mental health initiatives to identify the legal and ethical issues that have been raised. Most of the literature discussed privacy, mainly in terms of respecting the privacy of research participants. A small number of studies discussed ethics directly and indirectly, whereas legal issues were not substantively discussed in any studies, although some legal issues were discussed in passing, such as the rights of user subjects and privacy law compliance. From the papers that discussed ethics, the authors summarized the following ethical concepts: (1) *privacy*; (2) *security*; (3) *safety*; (4) *transparency*; (5) *autonomy*; and (6) *justice*.

On the topic of AI in mental health applications several reviews emphasize critical ethical concerns that must be addressed (Rubeis, 2022; Terra et al., 2023; Joseph & Babu, 2024; Gutierrez et al., 2024; Zhang & Wang, 2024; Thakkar et al., 2024). Key issues include *accountability* and *transparency*, where AI systems must offer interpretable decision-making processes to ensure trust among patients and clinicians; *privacy* and *data security*, highlighting the need for strong protections against unauthorized access to sensitive mental health data; *bias* and *equity*, as AI may perpetuate societal inequalities if trained on biased data, leading to unequal care; and *autonomy* and *informed consent*, ensuring that AI respects patient autonomy while balancing their needs with available resources. Additionally, concerns about the lack of empathy, human connection, and continuity of care in AI-driven systems stress the importance of human oversight to ensure AI supports, rather than replaces, human professionals.

## 3.2 Recommendation Checklists

Several contributions have highlighted that many apparently positive findings in machine learning applications were actually caused by methodological errors (Kapoor & Narayanan, 2023) that may occur in the experimental setup and in data collection (Paullada et al., 2021; Santy et al., 2023; Peng et al., 2021). Moreover, it is well-known that faulty data collection strategies may lead to biased machine learning models, whose application in real-life scenarios may have harmful implications on people (e.g., discrimination and stereotypes) (Birhane, 2022; Birhane et al., 2023).

Besides the risk that these factors pose to a machine learning model and, indirectly, to its consumers, researchers have pointed out the importance of thoroughly documenting their methodology to foster research reproducibility, awareness of technology consumption, and potential ethical issues. In particular, literature work has proposed recommendations and desiderata targeting these factors through standardized sheets, checklists, and '*how-to*' cards. Notable examples include datasheets for documenting and tracking data collection (Gebru et al., 2021; Pushkarna et al., 2022; Stoyanovich & Howe, 2019; Zhao et al., 2024b) and its derived software (Bender & Friedman, 2018), cards and factsheets for model selection (Mitchell et al., 2019; Adkins et al., 2022; Sokol & Flach, 2020; Arnold et al., 2019), ethics sheets for evaluating AI tasks (Mohammad, 2022) and their social impact (Shen et al., 2021; Derczynski et al., 2023), sheets for instructing on human evaluation (Shimorina &

Belz, 2022) and annotation (Díaz et al., 2022), and checklists for paper submission (Magnusson et al., 2023).

These tools aim to instill technology awareness in daily machine learning system practitioners and promote reproducibility. However, they are often meant as starting points for critical thinking, therefore they are intentionally broad and general-purpose and may not capture factors tailored to specific domains. For this reason, ad-hoc extensions of such recommendations have been proposed targeting healthcare datasets (Rostamzadeh et al., 2022) and ethical and data collection issues on speech-language technology (Papakyriakopoulos et al., 2023; Toussaint et al., 2022; Feng et al., 2022; Mahajan & Shaikh, 2021). In the domain of speech datasets concerning language impairment, Westerhout and Monachesi (2006) address aphasia providing a short list of requirements.[6]

To the best of our knowledge, we are the first to specifically address the domain of speech datasets for MHND.

## 4. Desiderata

Our analysis leads us to define a of list desiderata composed of: (i) informed consent, (ii) data storage and security, (iii) data privacy, (iv) accountability, (v) fairness, bias, and diversity, (vi) data quality, validation, and maintainability, (vii) discourse genre. Since the legal framework of reference will depend on the country where the data is collected, our list does not include normative obligations, which nevertheless should be properly taken into account by researchers. We shall remark that although we treat desiderata as separate concepts, there is often some overlap in the literature. For example, Fadda et al. (2022) discuss the topic of informing participants both about *users' privacy expectations and informed consent* and *data confidentiality.*

### 4.1 Informed Consent

In the context of clinical and health research environments, it is crucial for consent procedures to thoroughly communicate to individuals the specifics of when and how their data is being collected and utilized (Martinez-Martin et al., 2021).

In this domain, informed consent can be defined with the following three key components - information, comprehension, and voluntariness (Andrews et al., 2023). Participants, especially in sensitive areas like MHND, may not anticipate their speech being observed and analyzed for research purposes. Thus, securing informed consent ensures their understanding and agreement with the research parameters. This commitment extends beyond the initial consent to subsequent phases, where data sharing and reuse may occur, underlining the ethical responsibility to respect participants' autonomy.

Moreover, the collection, utilization, and sharing of speech data, particularly identifiable information, are contingent upon explicit and ongoing informed consent. Participants should retain the right to withdraw consent at any point, acknowledging the sensitive nature of MHND-related data.

---

6. Aphasia can be a symptom of a neurological disorder or brain injury, but is not considered a disorder in itself. For this reason, we do not address it in our work.

Transparency is pivotal in this domain. Comprehensive information concerning all the processes in which data will be involved, such as collection, storage, and analysis, should be communicated transparently. This includes the genres of speech data collected, potential inferences, intended usage, generated reports, recipients of data and reports, associated risks, benefits, and limitations.

Stakeholders, including participants, clinicians, and researchers, should have accessible explanations tailored to their specific interests and concerns. This comprehensive approach to consent and transparency fosters an ethical foundation for speech dataset utilization in MHND research.

## 4.2 Data Storage and Security

In previous research on mental health, the importance of data storage and security in digital contexts has been underscored. Recommendations include employing encryption and involving IT security teams to protect online activities (Fadda et al., 2022). Moreover, transparent communication about data handling practices, especially regarding high-risk scenarios, is essential (Fadda et al., 2022). Clear security standards and regular audits are crucial to monitor compliance and address vulnerabilities (Martinez-Martin et al., 2021). Additionally, the transition to digital platforms poses increased risks of data leakage and loss, raising concerns about privacy and trust in mental healthcare (Rubeis, 2022).

In light of these considerations, for MHND speech datasets, we advocate for stringent security measures. This includes ensuring secure storage through encryption and also addressing the scenario where data, while not openly accessible through open access, may still be retained and accessible upon request.

In instances where data are transferred through different devices, such as the collection of speech datasets through crowdsourcing, the focus on data security also extends to these transmissions. It is crucial to explicitly outline and adhere to secure transport encryption protocols, reinforcing a comprehensive approach to safeguarding sensitive data throughout the entire data collection process.

## 4.3 Data Privacy

Existing literature emphasizes the need for transparent regulations to address privacy concerns (Wykes et al., 2019), with granular consent proposed as an effective means to empower users in digital settings (Kim et al., 2017). Privacy, often intertwined with consent, emerges as a critical aspect in safeguarding users' data (Rubeis, 2022). The evolving landscape, encompassing digital phenotyping and various technologies, raises concerns regarding data protection, reidentification risks, and the sale of health inferences (Martinez-Martin et al., 2021). To mitigate these challenges, standards like differential privacy measures are recommended (Martinez-Martin et al., 2021). Notably, the intersection of social media, mobile apps, sensing technology, and chatbots in mental health applications reveals a focus on detection but a limited discussion on ethical implications (Gooding & Kariotis, 2021). When utilizing public online data, ensuring informed consent and respecting users' privacy expectations becomes imperative (Fadda et al., 2022).

Our approach involves a deliberate separation of the concepts of privacy and consent. When referring to privacy in our context, we encompass a spectrum of considerations es-sential for ensuring protection, especially when data are made open access or are accessible to third parties. This involves meticulous attention to factors such as anonymization and the management of non-identifiable aspects. This commitment to safeguarding privacy extends beyond the initial data collection phase, as it remains crucial when sharing data with collaborating partners. Notably, the responsibility for maintaining privacy measures should be ingrained in every step of the data lifecycle, from the physical collection of data to their management and eventual analysis. This holistic approach underscores our commitment to upholding privacy standards at every stage of the data handling process.

In light of the above, our desiderata include (i) careful avoidance of identifying participants as members of vulnerable groups, (ii) rigorous redaction of privacy-sensitive and identifying information, (iii) partial or full anonymization when sharing data with third parties.

These recommendations aim to foster responsible data practices, acknowledging the evolving landscape of MHND technologies while prioritizing user privacy and informed consent.

## 4.4 Accountability

Accountability in the context AI applications for mental health, emerges as a multifaceted concept, with its roots traced through various dimensions in existing literature.

First-wave accountability, observed in the development of mental health apps, concentrates on linguistic inclusivity. This involves ensuring that linguistic corpora of stimuli and responses adequately represent diverse communities with distinct ethnicity and modes of self-presentation (Gooding & Kariotis, 2021). Transitioning beyond linguistic considerations, second-wave concerns adopt a law and political economy approach. This perspective questions whether mental health apps might prematurely disrupt markets and potentially substitute expert professionals with more cost-effective, albeit limited, software. The second wave prompts critical inquiries into the beneficiaries and burdens associated with data collection, analysis, and use (Gooding & Kariotis, 2021).

In the development of digital phenotyping tools, ethical considerations stress the importance of an independent interdisciplinary review for potential ethical issues. This review process begins early in the development stages, ensuring responsible and ethical practices (Martinez-Martin et al., 2021). Moreover, in this regard, the imperative to adhere to current and future laws is a central ethical consideration (Andrews et al., 2023).

Insights from discussions on the role of AI in psychiatry further underscore the significance of accountability. While AI holds potential in assisting data analysis and predictions, it is recognized that it cannot fully replace the indispensable human elements of psychiatry, such as empathy and the therapeutic relationship between patients and psychiatrists (Terra et al., 2023). Responsible AI use in psychiatry requires a nuanced understanding of its limitations and ethical implications, emphasizing transparency and accountability (Terra et al., 2023). Against this backdrop, our desiderata for constructing speech datasets for MHND are: (i) conduct thorough ethical review and approval prior to data collection to align with established ethical standards; (ii) collect information on participants' country of residence to ensure dataset compliance with current and future laws, respecting regional ethical guidelines.

## 4.5 Fairness, Bias, and Diversity

Within the broader health landscape, the imperative to address bias, ensure fairness, and promote diversity pervades discussions that extend into the development of speech datasets for mental health (Martinez-Martin et al., 2021). The foundational principles of identifying and minimizing bias in digital phenotyping tools resonate deeply. Collaborative efforts and partnerships are emphasized to navigate potential biases in diverse communities and local contexts, highlighting the need for a collective commitment to fair representation.

Extending beyond digital phenotyping, the discourse on bias mitigation in algorithm development is paramount. This multifaceted approach involves scrutinizing training data, refining algorithmic processes, addressing biases in tool transfer across contexts, and ensuring an unbiased interpretation of digital phenotyping findings. Periodic reviews and re-evaluations are advocated, acknowledging the dynamic nature of bias management.

Diversity in research participants is underscored as a crucial component of equity and fairness. The lack of representation based on race, gender/sex, or disability in research data raises concerns about potential disparities in the development and implementation of digital phenotyping tools. The call for practices that account for the diversity of communities and contexts during design and development processes echoes the importance of mitigating harm to marginalized groups. It is evident from existing literature, particularly in computer science and MHND research, that the distinction between *sex* and *gender* is often conflated, despite their fundamental differences. In these studies, the term *gender* is frequently used to encompass both biological sex and socially constructed roles. Kaufman et al. (2023) are some of the latest from many authors who clarify this distinction, defining *sex* primarily as biological factors and *gender* as socially constructed roles.

It is worth noting, researchers' opinions about this terminology are complex and opposing, with some authors affirming that *sex* should be used when reporting biological differences between males and females, and *gender* for sociocultural differences between women and men (Muehlenhard & Peterson, 2011; Kaufman et al., 2023), but others arguing that the biological and sociocultural factors are closely intertwined, such that sex cannot be studied without considering gender (and vice versa), and thus the distinction between the terms sex and gender should be abandoned (Yoder, 1999; Van Anders, 2015). Following the suggestions of Hyde et al. (2019), in our paper we adopt the term *gender/sex*, both as the simplest way to discuss research which uses the two terms interchangeably, as well as a recognition of the complexity of the issue. We would however like to use this opportunity to reach out to the community and advocate for a more responsible use of the appropriate labels in this area in the future.

As ethical considerations unfold in constructing speech datasets, acknowledging representational and historical biases becomes indispensable. Diversity in speech datasets is multifaceted, encompassing (a) the distribution of languages, (b) varieties within a single language, and (c) speaker socioeconomic and demographic factors (Papakyriakopoulos et al., 2023). This diversity also involves various socioeconomic and demographic factors like nationality, gender/sex, age, education, income, and even health status. For instance, non-binary individuals and certain age groups, like young children, are often underrepresented (Papakyriakopoulos et al., 2023). It is important to recognize that while no dataset can fully capture all variations of human speech, it must reflect the specific domain of the intended application. For example, if the target application serves multiple ethnicities, the dataset must represent all these ethnic groups to ensure fair performance across populations. Recognizing that biases learned by machine learning models can perpetuate harmful stereotypes emphasizes the need for meticulous curation of datasets to avoid reinforcing societal prejudices (Andrews et al., 2023). The nuances of gender/sex, race, and age in dataset creation reveal instances of disparate algorithmic performance and underscore the imperative to address biases within these dimensions.

Furthermore, the digital divide and accessibility issues add another layer to the ethical discourse. Ensuring inclusivity beyond traditional attributes like age, gender/sex, and race, by considering other sensitive attributes and addressing accessibility concerns, emerges as a crucial facet. To counter confused taxonomies and enhance accuracy, self-reported annotations are proposed, empowering individuals to define their identity and minimizing misclassification. Additionally, providing fair treatment and compensation to contributors emphasizes the ethical responsibility towards those involved in dataset creation.

In the broader context of AI and Big Data in mental health, the implications of bias go beyond technical challenges, extending to potential discrimination and exclusion. The focus on social, gender/sex, and ethnic determinants in training data is crucial for avoiding biased AI systems and fostering a more personalized treatment approach (Rubeis, 2022).

These considerations converge into our desiderata: (i) ensure fair representation by developing effective strategies for identifying and minimizing bias in the development of speech datasets, ensuring fair representation such as balancing the distribution of demographics of the participants; (ii) identify and highlight limitations and biases that may be present in the data, discussing potential repercussions; (iii) implement self-reported annotations for diverse identity attributes, provide open-ended response options, acknowledge the multiplicity of identity, and collect aggregate data for commonly ignored sensitive attributes to enhance fairness and inclusivity in speech datasets for MHND; (iv) when conducting research, implement strategies to mitigate bias at different levels of algorithm development, spanning training data, algorithmic processes, tool transfer, and the interpretation of findings.

## 4.6 Data Quality, Validation, and Maintainability

In the realm of AI applications for MHND, the literature often lacks specific discussions on ethical aspects related to data quality, validation, and maintainability. In particular, there was consensus regarding the need to have a mechanism for reviewing or auditing the validity of digital phenotyping tools beyond their initial deployment, such as evaluating software updates or device uses deployed in new contexts (Martinez-Martin et al., 2021). However, while these standards primarily address the application viewpoint, there is a significant gap in considering the connected data. Recognizing this gap, we propose guidelines tailored for the construction of MHND speech datasets, synthesizing insights gleaned from an extensive literature review. First and foremost, we advocate for a meticulous verification of the speech data's quality to mitigate errors and inaccuracies. This involves ensuring that the dataset faithfully represents the relevant distribution of the intended application domain. In particular, a dataset should reflect the diversity of speech patterns, accents, speech disorders, and demographic factors relevant to the research objectives. This includes capturing a broad range of speech characteristics, such as those seen in individuals with mental health or neurological disorders, and accounting for variations in gender/sex, age, ethnicity, and socioeconomic background. By including such diversity, we reduce the risk of bias in speech recognition models, ensuring they perform more accurately across all demographic and linguistic groups. As discussed in Section 4.5, it is important to recognize that while no dataset can fully capture all variations of human speech, it must reflect the specific domain of the intended application.

Moreover, it is essential to describe the recording setting in detail, encompassing environmental specifics, technical aspects of recording instruments (such as sample rate and the presence of noise reduction systems), and the use of personal or environmental microphones. These technical parameters significantly influence data quality and validity, making detailed documentation imperative for researchers and clinicians to assess reliability and relevance.

Annotation of data, especially in the context of MHND, is underscored as a key step. In addition to standard annotation practices (e.g., the usage of self-reported questionnaires to retrieve PHQ-9 scores for depression), involving domain experts with a broad understanding of MHND nuances is crucial for contextualizing and interpreting the data accurately. Recent work suggest that when using self-report measures, the goal of the study must be reframed from predicting diagnosis to predicting self-report questionnaire scores, which may not always match clinical diagnosis (Low et al., 2020). Moreover, specifying the diagnostic manual (i.e., diagnostic criteria) used for reference, such as the DSM-IV or DSM-V for the case of depression, is important for at least two reasons: *clinical interpretation* and *research reproducibility*. Indeed, specifying the diagnostic criteria allows for better clinical interpretation of the findings. Clinicians need to know exactly what criteria were applied to understand the results in the context of their practice. Moreover, explicitly stating the diagnostic criteria ensures that other researchers can reproduce the study accurately. This is important for building upon existing research and validating findings in different populations or settings. For the sake of clarity and completeness, authors should always explicitly specify the diagnostic criteria, even when they could be inferred from other information (e.g., in the case of the PHQ-9 aligning with DSM-IV criteria).

Furthermore, maintainability stands as a pivotal factor in ensuring the reliability of data. In the realm of MHND speech datasets, the acknowledgment that modifications may occur underscores the need for a systematic approach to tracking these changes. This is essential for users who rely on clear and transparent references to specific dataset versions.

A resilient platform for dataset construction plays a central role in facilitating maintainability. This platform should not only support effective versioning, allowing authors to document and manage changes seamlessly, but also incorporate mechanisms for users to report issues. Accessibility is paramount, and a user-friendly platform significantly contributes to ensuring that users can easily locate and utilize the dataset.

Emphasizing accessibility goes hand in hand with fostering an environment where users can actively contribute to maintaining and improving the dataset. Therefore, a transparent approach to data maintenance involves not only making the dataset easily accessible but also establishing clear protocols for users to report issues and actively participate in ongoing improvements.

In summary, incorporating versioning and issue-tracking capabilities within the dataset platform is crucial for ensuring continuous usability and enhancement. This not only aligns with ethical considerations but also fortifies the overall robustness and reliability of MHND speech datasets in the ever-evolving landscape of AI applications.

## 4.7 Discourse Genre

In the development of speech datasets for MHND, an additional ethical consideration involves integrating the concept of *discourse genre*. This becomes especially pivotal when dealing with datasets associated with sensitive MHND contexts. *Discourse genre* encompasses the specific communication patterns and linguistic structures used in conversations, therapy sessions, or support interactions related to MHND. These genres vary based on the cognitive and linguistic influences imposed on a speaker (Bliss & Mccabe, 2006). For instance, therapeutic dialogue or narrative retelling are two examples of discourse genres, each requiring distinct linguistic analyses.

Our observation highlights that the interpretation of the term *discourse genre* fluctuates depending on the context of analysis. For instance, in datasets aimed at assessing language disorders, it is frequently denoted by the term itself (Bliss & Mccabe, 2006). In the context of aphasia *discourse genres* (Forbes et al., 2012) or *speech tasks* are commonly used. Bipolar disorder takes the form of *speech tasks* (Ciftçi et al., 2018), while in cases of depression, it pertains to *speaking patterns* (Cai et al., 2022) or *tasks* (Tasnim et al., 2022). In the context of dementia, it is linked to *stimuli tasks* (Guo et al., 2021), *discourse types* (Lanzi et al., 2023) or *language production contexts* (Kempler et al., 1987).

In response, we chose to adopt the term *discourse genre* because it serves as a convenient umbrella for referencing the diverse communication patterns used in constructing speech datasets for MHND. However, it is essential to note that in certain cases, such as content derived from online media, pinpointing a specific discourse genre may not be feasible.

In the context of speech MHND datasets, we propose a desideratum that underscores the importance of providing evidence of validity for the intended use of the data. This involves substantiating that the chosen discourse genres within the dataset are relevant and appropriate for the specified context (Low et al., 2020). For example, in aphasia analysis, narrative tasks may constitute a relevant discourse genre (Varlokosta et al., 2016; Forbes et al., 2012), while in depression research, the distinction between discourse genres like reading tasks and interview tasks could be significant (Cai et al., 2022; Tao et al., 2023).

It is essential to note that this desideratum specifically applies to datasets dealing with particularly sensitive MHND information. By establishing evidence of validity, dataset creators ensure that the discourse genres chosen are suitable for the intended context and align with ethical considerations for MHND data. This rigorous validation process helps in maintaining the integrity of the dataset, promoting responsible use in AI applications tailored for MHND support and analysis.

## 5. Methodology

In our investigation, we overview existing literature trying to encompass a broad spectrum of MHND as well as several methods to acquire speech data. Our study includes depression, anxiety,[7] dementia, Alzheimer's, bipolar disorder, somatic symptoms, stress, and Parkinson's. As data sources, we individuate personal interaction with participants, online media, and crowdsourcing. For what concerns the genre of the discourse, we include clinical interviews (both unstructured and semi-structured), spontaneous reading, spontaneous speech, speech tasks, and semantic fluency tests.[8]

## 5.1 Selection Criteria

We surveyed relevant literature following a non-systematic approach. We searched titles and abstracts using common keywords related to MHND, and speech datasets. First, we explored scientific venues that we consider related to the topic.[9] Then, we extended our search through the Google Scholar search engine. Finally, we examined datasets in the DementiaBank platform,[10] a well-known repository for datasets concerning MHND, and retrieved publications corresponding to the datasets we are interested in.

Inspired by Li et al. (2019), we then exclude any paper that does not meet all the following criteria: (1) they target patients with neurological disorders; (2) they record speeches by audio or video; (3) they function as public research resources for other researchers; (4) they are peer-reviewed publications. The 36 papers thus selected constitute the final corpus on which we will perform our analysis.

## 5.2 Checklist

To facilitate a structured analysis, we developed a checklist encompassing key considerations reflecting our desiderata. The following checklist is organized into distinct elements and serves as a guide to evaluate the presence or absence of best practices in the reviewed papers.

C1) **Informed Consent.** The authors provide information on sharing informed consent details with participants.

C2) **Data Storage and Security.** The authors report data storage and security measures.

C3) **Data Privacy.** The authors report information on data privacy measures such as data anonymization, removal of personal information, and naming conventions of data.

C4) **Accountability.** The authors report information about ethical board approval.

C5) **Fairness, Bias, and Diversity.** The authors provide information about data collection diversity and specific considerations for MHND data, including Alzheimer's and Parkinson's, in particular:

---

7. In literature, anxiety is often mentioned without clear delineation of whether it pertains to a symptom or a syndrome. Since our focus is on MHND, our attention is directed toward datasets that explicitly outline the assessment criteria or scales employed for evaluating anxiety. As a result, we omitted the dataset from Chollet et al. (2016) due to insufficient information provided regarding the diagnosis of anxiety. Regarding the DAIC-WoZ dataset (Gratch et al., 2014), it was included only for depression analysis since the available data contain only labels for this task.

8. For detailed definitions of each speech type, please refer to Appendix A.

9. Further details can be found in Appendix B.

10. https://dementia.talkbank.org/access/

C5-a  Speakers distribution per class.

C5-b  Age distribution of the speakers. Authors are required to disclose the average age of each participant group along with at least one additional statistic, such as standard deviation. Alternatively, they should provide the number of participants within specific age ranges, adhering to the inclusion criteria for participant selection.

C5-c  Gender/sex distribution of the speakers.

C5-d  Educational level of the speakers.

C5-e  Language of the speakers.

C5-f  Ethnicity of the speakers.

C5-g  **Parkinson's only**. Distribution of Unified Parkinson's Disease Rating Scale (UPDRS) and Hoehn and Yahr Scale (H&Y) among participants.[11]

C5-h  Biases awareness. The authors comment on the possible presence of bias in the dataset.

C5-i  Diversity collection measures. If the authors explicitly mention if they adopted diversity collection measures.

C6)  **Data Quality, Validation, and Maintainability.**

C6-a  The authors provide information about technical details on recording settings such as brand and version of the microphone, sampling frequency in Hz or kHz, loudness setting in dB, distance of the microphone from the speaker, and environmental conditions during recording.

C6-b  There is information on data maintainability, namely if the platform where data can be accessed allows versioning and issues reporting.

C6-c  *Clinical Domain.* There is information on annotation and validation by domain experts, such as the presence of domain experts in the loop.

C6-d  *Clinical Domain.* Authors specify the criteria used for diagnosis, regardless of whether it is done by a clinician or through self-assessment questionnaires.[12]

C6-e  *Clinical Domain.* **Alzheimer's only**. There are details about the cognitive-linguistic battery. Please refer to Appendix A for the definition of cognitive-linguistic battery.

---

11. Please, refer to Appendix A for details about UPDRS and H&Y scales.

12. When clinicians solely conduct diagnosis, we expect authors to furnish information akin to the following examples: "*In contrast to previous work, all participants met DSM-4 or DSM-5 criteria for major depression as determined by diagnostic interview*" (Dibeklioglu et al., 2018); "*The BDD patients were evaluated and diagnosed by an experienced psychiatrist. All of the patients satisfied both ICD-11 (International Classification of Diseases. 11th Revision) and DSM-5 (Diagnostic and Statistical Manual of Mental Disorders, Fifth Edition) criteria*" (Llamocca et al., 2021). When the diagnosis is performed through self-assessment questionnaires we expect authors to furnish information akin to the following examples: "*All MDD patients received a structured Mini-International Neuropsychiatric Interview (MINI) that met the diagnostic criteria for major depression of the Diagnostic and Statistical Manual of Mental Disorders (DSM) based on the DSM-IV*" (Cai et al., 2022); "*These questions are designed based on DSM-IV and other depression scales such as the Hamilton depression rating scale (HDRS)*" (Guo et al., 2021).

C7) ***Clinical Domain* - Discourse Genres.**

    C7-a The authors provide information about the discourse genres participants performed.

    C7-b The authors provide motivations for choosing specific discourse types.

## 6. Discussion

We present an overview of the 36 surveyed datasets in Table 1, detailing their discourse genre, the identified target issues, the sources of speech content, and illustrating their compliance with our checklist. For the sake of brevity, in the rest of this Section, we will abbreviate target issues and types of sources as explained in Appendix A.

Most datasets provide information about speakers and age distribution (C5-a: 28, C5-b: 25), the language of the speakers (C5-e: 24) and discourse genre performed by the participants (C7-a: 28). More than half of the selected works report information about informed consent (C1: 20), accountability (C4: 21), the gender distribution of the speakers (C5-c: 23), presence of domain experts in the loop (C6-c: 17) and reasons for selecting specific discourse genres (C7-b: 17). However, the remaining items are less prominently addressed.

This result indicates a general awareness among the researchers of the importance of reporting the speakers' distribution across different dimensions. However, the speakers' ethnicity is seldom reported. About half of the examined papers do not report on informed consent and accountability. This could be problematic, for a number of reasons. First of all, it suggests that these aspects are not sufficiently emphasized in the community. Moreover, it undermines the stability of datasets in case future data policies prevent the use of these data. Likewise, the low coverage in data storage and security (C2: 9) and data privacy (C3: 10) suggests a tendency to overlook measures for the safeguard of participants' sensitive information. Finally, we observe low coverage in data quality, validation, and maintainability (C6-a: 12, C6-b: 8, C6-d: 8), which could hamper the widespread utilization of this data.

In the remainder of this section, we analyze our data along three dimensions: (i) the target issues (ii) the discourse genre and (iii) the sources contributing to the dataset's content. The first and second dimensions aim to chart the community effort concerning specific target issues and discourse genres. The third one intends to highlight potential systematic shortcomings of specific speech sources. For example, datasets created through crowdsourcing efforts may systematically lack information about recording settings due to the inherent constraints of the collection method. Understanding such patterns is crucial in identifying potential pitfalls of specific content sources.

| Dataset | Discourse Genre | Target Issue | Source | C1 | C2 | C3 | C4 | C5-a | C5-b | C5-c | C5-d | C5-e | C5-f | C5-g | C5-h | C5-i | C6-a | C6-b | C6-c | C6-d | C6-e | C7-a | C7-b |
|---|---|---|---|---|---|---|---|---|---|---|---|---|---|---|---|---|---|---|---|---|---|---|---|
| Zou et al. (2023) | CI | DP | PI | ● | | ● | ● | ● | ● | ● | | ● | | | ● | | | | | | | ● | |
| Tlachac et al. (2021) | SR, SS | DP, AX | PI | ● | ● | ● | ● | | | | | | | | | ● | | | | | - | ● | ● |
| Dogrucu et al. (2020) | SR | DP | CS | ● | ● | | | | | | | | | | | | | | | | - | | |
| Harati et al. (2020) | CI | DP | PI | | | | ● | ● | | | | ● | | | | | | | ● | | - | ● | |
| Lin et al. (2023) | CI | DP, AX | PI | ● | | | | ● | ● | ● | | ● | | | | | | ● | ● | | | | ● |
| Tlachac et al. (2022) | CI | DP | CS | ● | ● | ● | ● | ● | ● | ● | ● | ● | | | ● | ● | | | | | - | ● | |
| Tao et al. (2023) | CI, SR | DP | PI | | ● | ● | ● | ● | ● | ● | | ● | | | ● | | | | ● | ● | - | ● | ● |
| Cai et al. (2022) | CI | DP | PI | ● | ● | ● | ● | ● | ● | ● | | ● | | | | | ● | | ● | ● | | ● | ● |
| Song et al. (2023) | SS | DP, AX | CS | | | | | ● | ● | ● | | ● | | | ● | | | | | | - | | |
| Tasnim et al. (2022) | ST | DP, AX | CS | | | | ● | ● | ● | ● | ● | ● | | | ● | | | | ● | ● | | ● | ● |
| Gratch et al. (2014) | CI | DP | PI | ● | | ● | | | | | | ● | | | | | | | | | | | |
| Shen et al. (2022) | CI | DP | PI | ● | | | | | ● | | | | | | ● | | | ● | | | - | ● | |
| Dibeklioglu et al. (2018) | CI | DP | PI | | | | | | ● | | | ● | | | | | | | ● | ● | | ● | |
| Guo et al. (2021) | CI | DP | PI | ● | | | ● | ● | ● | ● | | ● | | | | | | ● | ● | | - | ● | |
| Alghowinem et al. (2018) | CI | DP | PI | | | | ● | ● | ● | | | ● | | | ● | | | | | | | ● | |
| Yoon et al. (2022) | SS | DP | OM | | | ● | ● | ● | | | ● | | | | | | | | | | - | | |
| Kempler et al. (1987) | SS, ST | DM | PI | | | | | ● | ● | | ● | | | | | | | | ● | | | | |
| Kurtz et al. (2023) | ST | DM | PI | | | | | ● | ● | ● | | | | | ● | | | | | | - | | |
| Lanzi et al. (2023) | SR | AL | PI | ● | ● | ● | | ● | ● | ● | ● | ● | | | ● | ● | | | ● | | | ● | |
| Luz et al. (2021) | SS | AL | PI | | | | ● | ● | ● | ● | | ● | | ● | | | | | | | | | |
| Ivanova et al. (2022) | SR | AL | PI | ● | | | ● | ● | ● | ● | ● | | | | | | | ● | ● | | | | |
| Novikova and Balagopalan (2020) | SS | AL | PI | | | | | ● | | | | | | | | | | | | | | | |
| Becker (1994) | CI | AL | PI | ● | | | | ● | ● | ● | | ● | | | | | | | ● | | | | |
| Ciftci et al. (2018) | CI | BD | PI | | | | | ● | ● | ● | ● | ● | | | | | | | ● | ● | - | ● | |
| Wang et al. (2020) | CI | BD | PI | ● | | | | ● | ● | ● | | ● | | | | | | | ● | ● | | | ● |
| Llamocca et al. (2021) | CI | BD | PI | | | ● | ● | | | | | ● | | | | | | | ● | | | | |
| Gideon et al. (2016) | CI, SS | BD | PI | | ● | | ● | ● | ● | | | | | | | | | ● | | | - | ● | |
| Qian et al. (2023) | ST | SSD | PI | ● | ● | | ● | ● | ● | ● | | | | | ● | | | | | | - | ● | |
| Chaptoukaev et al. (2023) | ST | S | PI | ● | ● | ● | ● | | ● | ● | | ● | | | ● | | | | ● | | | ● | ● |
| Jaiswal et al. (2020) | ST | S | PI | | | | | | | ● | ● | ● | | | | | | | ● | | - | ● | |
| Hansen and Bou-Ghazale (1997) | CI, ST | S | PI | | | | | | | ● | ● | | | | | | | | | | | | |
| Chen et al. (2022) | ST | S | PI | | | ● | ● | | | | | ● | | | | | | | | | - | ● | |
| Zhang et al. (2022) | SFT | P | PI | ● | | | ● | | ● | ● | | ● | | | | | | | | | - | ● | |
| Sakar et al. (2013) | SR | P | PI | | | | ● | | ● | | | ● | | ● | | | | | ● | | - | ● | |
| Orozco-Arroyave et al. (2014) | SR, SS, ST | P | PI | ● | | | ● | | ● | ● | | | | | | | | | ● | ● | - | ● | |
| Correia et al. (2018) | SS | P, DP | OM | | | | | | | | | ● | | | | | | | | | - | | |
| **Total** | 36 | 36 | 36 | 20 | 9 | 10 | 21 | 28 | 25 | 23 | 12 | 24 | 2 | 2 | 13 | 7 | 12 | 8 | 17 | 8 | 1 | 28 | 17 |

Table 1: Checklist for Mental and Neurological Health Speech Datasets.

|  | DP | AL | AX | BD | P | S | DM | SSD |
|---|---|---|---|---|---|---|---|---|
| **No. Datasets** | 17 | 5 | 4 | 4 | 4 | 4 | 2 | 1 |
| **C1** | 11 | 3 | 2 | 2 | 2 | 1 | 0 | 1 |
| **C2** | 4 | 2 | 1 | 1 | 0 | 1 | 0 | 1 |
| **C3** | 7 | 0 | 1 | 1 | 0 | 2 | 0 | 0 |
| **C4** | 9 | 3 | 2 | 3 | 3 | 2 | 0 | 1 |
| **C5-a** | 13 | 5 | 3 | 4 | 3 | 0 | 2 | 1 |
| **C5-b** | 11 | 4 | 3 | 2 | 3 | 2 | 2 | 1 |
| **C5-c** | 9 | 4 | 3 | 2 | 3 | 3 | 1 | 1 |
| **C5-d** | 4 | 3 | 1 | 1 | 1 | 2 | 1 | 0 |
| **C5-e** | 14 | 2 | 3 | 2 | 3 | 3 | 0 | 1 |
| **C5-f** | 1 | 1 | 0 | 0 | 0 | 0 | 0 | 0 |
| **C5-g** | - | - | - | - | 2 | - | - | - |
| **C5-h** | 7 | 1 | 2 | 1 | 1 | 2 | 1 | 1 |
| **C5-i** | 6 | 1 | 4 | 0 | 0 | 0 | 0 | 0 |
| **C6-a** | 4 | 2 | 0 | 1 | 2 | 0 | 0 | 1 |
| **C6-b** | 6 | 2 | 1 | 0 | 0 | 1 | 0 | 0 |
| **C6-c** | 8 | 2 | 1 | 4 | 2 | 0 | 1 | 0 |
| **C6-d** | 5 | 0 | 0 | 2 | 0 | 0 | 0 | 0 |
| **C6-e** | - | 1 | - | - | - | - | - | - |
| **C7-a** | 13 | 3 | 3 | 2 | 3 | 4 | 2 | 1 |
| **C7-b** | 8 | 3 | 2 | 1 | 2 | 2 | 1 | 0 |

Table 2: Number of datasets that address each checklist item grouped by target issue.

## 6.1 Dimension 1: Target Issue

In Table 2, we group the datasets according to their target issues to provide an overview of their compliance with the items in our checklist. Datasets encompassing multiple target issues are considered multiple times, once per issue. In our analysis, we will follow the same order represented in the table.

### 6.1.1 Depression

Among the 17 datasets concerning depression (**DP**), we observe that 11 incorporate details on informed consent (C1), 4 implement robust data storage and security measures (C2), 7 disclose information on data privacy measures (C3), and 9 report about accountability (C4). Regarding fairness, bias, and diversity aspects, we found that 13 datasets provide information on the distribution of speakers by class (C5-a), 11 inform regarding the age distribution of speakers (C5-b), 9 present the gender/sex distribution of speakers (C5-c), and 4 outline the education level of speakers (C5-d). Moreover, 14 datasets include information on different languages of speakers (C5-e), but only one elaborates on distinct speaker ethnicity for a specific language (C5-f). Lastly, 7 datasets exhibit awareness of potential bias in the dataset (C5-h), and 6 incorporate diversity collection measures (C5-i). Regarding aspects of data quality, validation, and maintainability, only 4 datasets provide information about the recording setting (C6-a), and 6 adhere to the principles of maintainability (C6-b). 8 datasets involve the presence of domain experts in the loop (C6-c), while the others automatically validate observations based on scores obtained from participants in depression self-assessment questionnaires. Among those involving domain experts, 5 out of 8 specify the type of clinical diagnosis manual adopted (C6-d). Finally, 13 datasets provide information regarding the speech tasks required of participants (C7-a), and only 8 of them justify that choice (C7-b).

We argue that the limited awareness of potential biases in the datasets and the relatively low adoption of diversity collection measures (C5) should raise concerns about the inclusivity and representativeness of the research. Moreover, the frequent absence of experts in the loop highlights a potential oversight, risking a lack of nuanced understanding and interpretation. The limited information on recording settings and adherence to transparency and maintainability principles further suggests a need for the scientific community to identify best practices in ensuring data quality (C6). The uneven distribution in providing justifications for the choices made about discourse genre suggests a need for more thorough and transparent methodological reporting. In conclusion, we believe that further reflection is needed on the integration of ethical considerations in depression datasets.

### 6.1.2 ALZHEIMER'S

According to Table 2 (column **AL**), the majority of datasets provide information on the distribution of speakers by class (C5-a), details about speaker age (C5-b) and speaker gender/sex (C5-c). Furthermore, around half of the datasets cover informed consent (C1), accountability (C4), information about speaker educational levels (C5-d), and discourse genre (C7-a and C7-b). However, only two datasets feature aspects such as data security (C2), language of the speakers (C5-e) and data quality, validation, and maintainability (C6-a, C6-b, C6-c). Similarly, a single dataset provides documentation regarding different ethnicities of speakers for a specific language (C5-f), bias awareness (C5-h), diversity collection measures (C5-i), and cognitive-linguistic batteries (C6-e). Notably, there is no mention of the adoption of data privacy measures (C3).

Overall, the analysis of Alzheimer's datasets reveals consistent attention to certain ethical dimensions, such as speaker distribution and demographic details. However, critical aspects like data security and privacy, as well as measures addressing bias and diversity, demand increased consideration in future research.

### 6.1.3 ANXIETY

From Table 2 (column **AX**), we observe that aspects such as informed consent (C1), speaker distribution by class (C5-a), speaker age (C5-b), speaker gender/sex (C5-c), different languages of speakers (C5-e), and information on discourse genres (C7-a, C7-b) are well-represented, with more than half of datasets reporting information in these areas. Moreover, all datasets report information about the adoption of diversity collection measures (C5-i).

However, other critical ethical dimensions such as data storage and security (C2), data privacy (C3), accountability (C4), educational level of participants (C5-d), and different ethnicity of speakers for a specific language (C5-f) are less frequently addressed, being present in at most two out of the seven collected datasets. Lastly, all aspects concerning data validity and maintainability (C6) are scarcely reported or entirely omitted.

In conclusion, while certain ethical considerations receive ample attention in anxiety disorder datasets, there remains a significant need for increased emphasis on crucial aspects such as data security, privacy, data quality, maintainability, and the involvement of domain experts to ensure a more robust ethical framework across the analyzed papers.

### 6.1.4 Bipolar Disoder

We collect 4 datasets addressing Bipolar Disorder (**BD**). We observe that the majority of them provide details on accountability (C4), speaker distribution (C5-a) and involve domain experts for data validation/annotation (C6-c). However, only half of these datasets include information on the clinical diagnosis reference manual (C6-d). Furthermore, aspects such as data storage and security (C2), data privacy (C3), speaker education level (C5-d), and bias awareness (C5-h) are scarcely addressed. Moreover, when it comes to data quality, only one dataset includes information about the recording setting (C6-a). Lastly, crucial aspects like the inclusion of speakers of different languages or the acknowledgment of different ethnicities in the dataset (C5-e, C5-f), the adoption of diversity collection measures (C5-i), and maintainability are not documented.

In conclusion, while some datasets provide valuable insights into speaker distribution and involve domain experts, there is a conspicuous lack of documentation on essential ethical considerations and aspects related to data quality.

### 6.1.5 Parkinson's

We collect 4 datasets related to Parkinson's (P). Three of them provide information on accountability (C4), as well as the distribution of speakers by class, age, gender/sex, and education level (C5-a, C5-b, C5-c) and discourse genres performed by speakers (C7-a). However, only two datasets offer justifications for their choice of discourse genres. Additionally, only half of the analyzed papers report details on informed consent (C1), the language of the speakers (C5-e) UPDRS and H&Y rating scales (C5-g), recording settings (C6-a), and validation by domain experts (C6-c). In contrast, only one dataset contains information on the education level of speakers (C5-d) or bias awareness (C5-h). Notably, none of the papers discuss data storage and security (C2), data privacy (C3), ethnicity of speakers (C5-d and C5-f), diversity collection measures (C5-i), or maintainability (C6-b).

Despite the limited availability of Parkinson's datasets for analysis, our results suggest that there are notable gaps in addressing crucial ethical considerations. Future research in Parkinson's studies should prioritize aspects like data security, privacy, diversity measures, and maintainability to ensure a more comprehensive and ethically robust framework for data collection in the context of Parkinson's disease. Moreover, we argue that more attention has to be paid to report recording setting details and UPDRS and H&Y rating scales.

### 6.1.6 Stress

Table 2 (column **S**) reports statistics about datasets targeting Stress. All analyzed datasets provide information on the discourse genre performed by speakers (C7-a). About half of them include details on the age and gender/sex distribution of speakers (C5-b, C5-c), measures of data privacy (C3), accountability (C4), level of speaker education (C5-d), biases awareness (C5-h), and rationale for choosing discourse genres (C7-b). Nonetheless, only one dataset includes information on crucial aspects such as informed consent (C1) and data storage and security (C2). Furthermore, no dataset addresses speaker distribution by class (C5-a), speaker's ethnicity (C5-f), and diversity collection measures (C5-i). Lastly, the majority of collected datasets do not report details about data quality, validation and maintainability (C6).

Overall, there is a general lack of information on key ethical considerations. In particular, aspects like data storage security, speaker distribution by class, and diversity collection measures are insufficiently addressed.

### 6.1.7 Somatic Symptoms Disorder and Dementia

A relatively low amount of datasets address Somatic Symptoms Disorder (SSM) and Dementia (DM). We observe a notable lack of information regarding informed consent (C1), data storage and security (C2), data privacy (C3), and accountability (C4). In the context of collection diversity and documentation tailored to address mental health diseases, we observe a good coverage of speakers distribution (C5-a) and their age (C5-b). Nonetheless, other relevant speakers' characteristics, like gender/sex distribution (C5-c), education level (C5-d), language (C5-e), ethnicity (C5-f), and diversity collection measures (C5-i), are scarcely addressed. Likewise, almost no dataset reports information concerning data quality, validation, and maintainability (C6)

Overall, despite the relatively small sample of datasets, our observations highlight a lack of documentation regarding the data collection process, the diversity of involved speakers, and, most importantly, its inherent quality and validation.

## 6.2 Dimension 2: Discourse Genre

Table 3 reports the number of datasets that employ specific discourse genres and provides a comprehensive overview of compliance with the items in our checklist. It is worth noting that some datasets encompass multiple discourse genres. Thus, the same dataset may appear multiple times when counting occurrences.

### 6.2.1 Clinical Interviews

A clinical interview serves as a fundamental discourse genre within the domain of healthcare, facilitating an in-depth interaction between patients and clinicians. In essence, it is a structured or semi-structured conversation designed to gather comprehensive information about a patient's medical history, current symptoms, and psychosocial factors. Unlike casual conversations, clinical interviews are purposeful and goal-oriented, aiming to elicit specific information crucial for diagnosis, treatment planning, and therapeutic interventions.

In our study, Clinical Interviews (**CI**) are the most prevalent discourse genre. A good portion of the datasets report informed consent (C1), highlighting a strong commitment to participant autonomy and transparency in research procedures. Additionally, there is significant attention given to fairness, bias, and diversity (C5), indicating a concerted effort to mitigate potential biases and ensure equitable representation within the datasets. However, only two datasets provide considerations about diversity collection measures (C5-i).

| | CI | ST | SS | SR | SFT |
|---|---|---|---|---|---|
| **No. Datasets** | 17 | 9 | 9 | 7 | 1 |
| **C1** | 12 | 3 | 2 | 6 | 1 |
| **C2** | 3 | 2 | 3 | 3 | 0 |
| **C3** | 6 | 2 | 2 | 2 | 0 |
| **C4** | 10 | 5 | 4 | 6 | 1 |
| **C5-a** | 15 | 5 | 7 | 5 | 1 |
| **C5-b** | 13 | 7 | 4 | 5 | 1 |
| **C5-c** | 10 | 7 | 4 | 5 | 1 |
| **C5-d** | 5 | 4 | 1 | 3 | 1 |
| **C5-e** | 14 | 6 | 3 | 3 | 1 |
| **C5-f** | 1 | 0 | 0 | 1 | 0 |
| **C5-g** | 0 | 1 | 1 | 1 | 1 |
| **C5-h** | 7 | 5 | 2 | 2 | 0 |
| **C5-i** | 2 | 1 | 2 | 3 | 0 |
| **C6-a** | 5 | 4 | 2 | 4 | 0 |
| **C6-b** | 5 | 0 | 2 | 3 | 0 |
| **C6-c** | 12 | 3 | 3 | 4 | 0 |
| **C6-d** | 7 | 1 | 0 | 1 | 0 |
| **C6-e** | 0 | 0 | 0 | 1 | 0 |
| **C7-a** | 14 | 9 | 5 | 6 | 1 |
| **C7-b** | 7 | 5 | 4 | 6 | 0 |

Table 3: Number of datasets that address each checklist item grouped by discourse genre.

Accountability (C4) emerges as another important aspect, demonstrating adherence to institutional review board protocols and ethical oversight. Moreover, it is worth noting that clinical domain-related items, such as the presence of domain experts in the loop (C6-c), and reporting discourse genre information (C7-a), are present in more than half of the datasets, even if information on clinical diagnosis is present only in seven datasets (C6-d).

Given the direct relationship between clinicians and patients during data collection through clinical interviews, it is reassuring to observe significant attention toward informed consent and accountability. However, we argue that greater attention should be paid to data storage and security as well as data quality, validation, and maintainability to mitigate risks and ensure the protection of sensitive participant information. Lastly, given the intimate nature of these interactions, where participants often divulge personal and sensitive information to clinicians, it becomes imperative to prioritize data protection and anonymization, especially when utilizing and sharing it for research endeavors.

### 6.2.2 Speech Tasks

Speech Tasks (**ST**) can encompass a broad spectrum of activities, ranging from simple repetition of phonemes to complex narrative retellings. The nature of these tasks is tailored to the objectives of the study, aiming to elicit specific linguistic features or behaviors indicative of underlying cognitive or physiological conditions.

As we can observe in Table 3, these datasets exhibit similar ethical considerations to CI datasets, albeit with some distinctions. While fairness, bias, and diversity considerations feature prominently (C5-b, C5-c, and C5-e), suggesting efforts to promote inclusivity and mitigate biases in dataset composition, there is a lack of documentation concerning informed consent (C1), accountability (C4), and biases awareness (C5-h). However, clinical domain-related items are less emphasized in the context of data validation (C6).

It is encouraging that all the analyzed datasets describe the speech tasks performed by participants (C7-a), even if around half of them also provide reasons for choosing them (C7-b). Unfortunately, there is scarce attention towards data storage and security (C2), diversity collection measures (C5-i), and data quality and maintainability (C6-a, C6-b). The aspect of privacy (C3) is often not addressed properly too, but it is important to consider that in this context participants would probably be less inclined to divulge sensitive information.

### 6.2.3 Spontaneous Speech

Spontaneous speech (**SS**) addresses the unscripted, fluid expression of language that occurs in everyday interactions. Unlike rehearsed or scripted speech, spontaneous speech tasks prompt individuals to articulate their thoughts, emotions, and experiences spontaneously, mirroring the dynamic nature of real-world conversations. Whether engaging in casual conversations, narrating personal anecdotes, or responding to open-ended prompts, participants generate spontaneous speech that reflects their linguistic abilities, cognitive processes, and sociolinguistic awareness.

Table 3 highlights that nearly half of the datasets provide details about accountability (C4), the distribution of participants by age and gender/sex (C5-b, C5-c), speaker interaction and corresponding motivation of such a choice (C7-a, C7-b). Additionally, a low portion of the datasets addresses data storage and security (C2), privacy (C3), educational level (C5-d), language (C5-e), ethnicity (C5-f), biases awareness (C5-h), and diversity collection measures (C5-i). Furthermore, there is a lack of focus on data quality, validation, and maintainability (C6). Lastly, it is particularly concerning the neglect of data privacy (anonymization), despite its significance in a scenario where people speaking spontaneously may report personal and sensitive information.

### 6.2.4 Spontaneous Reading

Spontaneous reading (**SR**) targets the natural flow and comprehension of written text among individuals. Unlike scripted readings or recitations, spontaneous reading tasks entail participants engaging with text material without prior preparation or rehearsal.

From Table 3 we observe that nearly all datasets address informed consent (C1) and accountability (C4). Furthermore, the majority of the datasets address fairness, bias, and diversity (C5-a, C5-b, C5-c) and data quality, validation, and maintainability (C6-a, C6-b, C6-c) Additionally, most datasets detail the types of spontaneous reading tasks assigned to participants and provide the rationale for their selection (C7-a, C7-b), a crucial aspect for understanding study objectives and participant characteristics relevant to syndrome detection or classification. However, few works discuss data storage and security (C2) or data privacy (C3), although participants may be less inclined to divulge sensitive information in this context. While participants may not disclose sensitive information, vigilance in these areas is crucial for maintaining data integrity and reliability. Similarly, information regarding the educational level (C5-d) and language (C5-e) of participants is lacking, despite its relevance to interpreting results from reading tasks fairly. Moreover, few datasets provide details about participant ethnicity (C5-f), bias awareness (C5-h), and diversity collection measures (C5-i).

### 6.2.5 Semantic Fluency Tests

Semantic fluency tests (**SFT**) evaluate individuals' cognitive functioning like their semantic memory and language abilities. The primary objective is to measure the individual's capacity to access and retrieve information from their semantic memory stores spontaneously. In these tests, participants are typically asked to generate words belonging to a specific semantic category within a limited timeframe, such as naming animals or fruits (Zhang et al., 2022).

In our survey, SFT is limited to the dataset provided by (Zhang et al., 2022). Drawing definite about the comprehensive coverage of aspects outlined in our checklist on a single dataset is challenging. However, we can infer that this dataset prioritizes certain aspects. Specifically, attention is given to informed consent (C1), accountability (C4), fairness, bias, and diversity considerations (C5). Additionally, the dataset includes a description of the discourse genre (C7-a). Nonetheless, no information about data evaluation, validity, and maintainability (C6) is reported.

## 6.3 Dimension 3: Source of Content

We observe that data is typically collected either via personal interactions with participants (PI), indirect methods such as crowdsourcing (CS), or by annotating media already available online (OM). We posit that CS or OM data collection prevents compliance with specific aspects of the checklist, whereas PI does not. Accordingly, in the analysis that follows, we focus on CS and OM. Table 3 offers a detailed view.

### 6.3.1 Crowdsourcing

Crowdsourcing (**CS**) involves outsourcing content-production tasks from a distributed pool of participants, who are tasked to engage in various activities, such as completing surveys, performing cognitive tasks, collecting social media and/or sensor data or providing subjective feedback on their health status.

In our survey, we consider 5 datasets gathered through crowdsourcing. We find it noteworthy that all of these datasets incorporate diversity collection measures (C5-i). This implementation is particularly seamless in crowdsourced data collection, as participants typically provide personal information themselves.

Information concerning recording settings (C6-a) is absent in all the papers. This is understandable, considering that in crowdsourcing participants often utilize their personal laptops or mobile devices, resulting in very diverse settings. In most cases, details about the involvement of domain experts and clinical diagnoses (C6-c, C6-d) are lacking. This suggests that data annotation is largely automated, possibly relying on self-assessment scores.

|  | CS | OM |
|---|---|---|
| **No. Datasets** | 5 | 2 |
| **C1** | 3 | 0 |
| **C2** | 3 | 0 |
| **C3** | 2 | 1 |
| **C4** | 3 | 0 |
| **C5-a** | 3 | 1 |
| **C5-b** | 3 | 0 |
| **C5-c** | 2 | 1 |
| **C5-d** | 2 | 0 |
| **C5-e** | 3 | 1 |
| **C5-f** | 1 | 0 |
| **C5-g** | 0 | 0 |
| **C5-h** | 2 | 0 |
| **C5-i** | 5 | 0 |
| **C6-a** | 0 | 0 |
| **C6-b** | 2 | 0 |
| **C6-c** | 1 | 0 |
| **C6-d** | 1 | 0 |
| **C6-e** | 0 | 0 |
| **C7-a** | 3 | 0 |
| **C7-b** | 2 | 0 |

Table 4: Datasets addressing each checklist item grouped by data collection method.

### 6.3.2 Online Media Annotation

Two of the datasets we analyzed concerned cases where the source videos and audios were already publicly available on YouTube. The authors of the datasets scraped such material, annotated it, and redistributed it.

In these cases, information regarding informed consent (C1), data storage and security (C2), and accountability (C3) are not provided. The point about data storage and security is understandable since the user content is not produced or managed by the authors of the dataset. However, we believe the issue of informed consent and accountability deserves attention. We are afraid that this is an inherent limitation of online media annotation, because collecting informed consent from the original content creators or the approval of ethical boards could be challenging if not outright unfeasible. Likewise, for the same reason, diversity collection measures (C5-i) are not reported in either case.

## 7. Conclusions

Our survey identified key areas for improvement in mental health and neurological disorders resource creation and use. About half of the examined papers do not report on informed consent and accountability. Other crucial aspects such as data storage, privacy, quality, validation, and maintainability have even lower coverage. These limitations are not always distributed equally among the domains of interest. For instance, the absence of validation from experts in the field predominantly impacts datasets related to Anxiety and Stress, as well as those gathered through discourse genres outside of clinical interviews. Another example concerns datasets collected through online media annotations, which typically lack information on informed consent, data storage and security, as well as accountability measures.

On a positive note, our survey has also demonstrated a rich and varied set of valuable resources for MHND research. Such resources are pivotal in driving progress toward reliable and trustworthy AI systems with a positive impact on society, and this survey may serve as a reference to the novice embarking on such an endeavor. However, we believe it is crucial that such resources are used with increased awareness of potential pitfalls. To this end, we compiled a list of desiderata and a checklist for collecting and documenting datasets for mental health and neurological disorders. The checklist is intended not only as a tool to analyze existing literature but also, perhaps more importantly, as a guideline for a more principled approach to future resource design, implementation, and maintainance and use in this field.

We are aware that the application of such guidelines may pose some challenges. One of them would be addressing the trade-offs between separate items, such as data privacy and data access (Bak et al., 2022). In this regard, researchers agree that each study needs to consider the benefits of both items at tension with each other, and find an equilibrium - e.g., balance the need for seamless exchange of information for improved healthcare outcomes with the strong obligation to protect the patient's privacy (Chitta et al., 2019); or balance the risks of a user's privacy loss with the benefits of delivering them a tailored experiences and targeted marketing from shared data (Malik, 2024). Even more so, in light of these challenges, we wish for our work to stimulate a closer collaboration between computer science and clinical domains for a sounder and more mutually beneficial resource development.

## Acknowledgments

We thank psychiatrist Fabrizio Mancini, MD, for his valuable insights and suggestions. This work was partially supported by the following projects: European Commission's NextGeneration EU programme, PNRR – M4C2 – Investimento 1.3, Partenariato Esteso, PE00000013 - "FAIR - Future Artificial Intelligence Research" – Spoke 8 "Pervasive AI"; EU H2020 ICT48 project "Humane AI Net" under contract #952026.

## Appendix A. Terminology

### A.1 Target Issue

- **Depression (DP).** A mood disorder causing a persistent feeling of sadness and loss of interest.

- **Alzheimer's (AL).** A progressive neurological disorder that causes memory loss and cognitive decline.

- **Bipolar Disorder (BD).** A mental illness characterized by extreme mood swings, including episodes of mania and depression.

- **Anxiety (AX).** A feeling of worry, nervousness, or unease, typically about an imminent event or something with an uncertain outcome.

- **Stress (S).** Feelings of emotional or physical tension, often in response to challenging or demanding situations.

- **Dementia (DM).** A decline in cognitive function severe enough to interfere with daily life and activities.

- **Somatic Symptoms Disorder (SSD).** A mental health condition characterized by physical symptoms that mimic physical illness or injury, but without a clear medical cause.

- **Parkinson's (P).** A progressive nervous system disorder affecting movement, causing tremors, stiffness, and difficulty with balance and coordination.

### A.2 Discourse Genre

- **Clinical Interviews (CI).** Conversational sessions conducted to gather information about an individual's mental health. These sessions are often either unstructured or semi-structured and may or not be conducted by a clinician. For example, in the case of crowdsourcing data collection, the sessions may not necessarily be conducted by a clinician.

- **Speech Tasks (ST).** Structured activities designed to assess various aspects of speech production, comprehension, or fluency.

- **Spontaneous Reading (SR).** Assessing reading abilities by asking individuals to read aloud a passage or text without prior preparation.

- **Spontaneous Speech (SS).** Assessing speech production and fluency by asking individuals to speak freely on a given topic without prior preparation.

- **Semantic Fluency Tests (SFT).** Assessments that measure an individual's ability to produce words belonging to a specific category within a limited time frame, often used to evaluate language and cognitive function.

**A.3 Source of Content**

- **Personal Interactions with the Participants (PI).** It refers to the direct engagement and communication between MHND professionals and participants.

- **Online Media (OM).** Audio-visual media uploaded on various online platforms, including but not limited to YouTube, or other vlogging platforms, which can encompass a wide range of content such as personal vlogs, educational material, entertainment, or any other form of audio-visual content available on the internet.

- **Crowdsourcing (CS).** The practice of obtaining information, ideas, or services by soliciting contributions from a large group of people, typically facilitated through online platforms or communities.

**A.4 Other**

- **MHND**: Mental Health and Neurological Disorders.

- **Unified Parkinson's Disease Rating Scale (UPDRS).** A comprehensive tool used to assess the severity and progression of Parkinson's disease by evaluating various aspects including motor function, activities of daily living, and mood.

- **Hoehn and Yahr Scale (H&Y).** A staging system used to measure the progression of Parkinson's disease based on the severity of motor symptoms and their impact on daily activities. It ranges from stages 1 to 5, with higher stages indicating more severe impairment.

## Appendix B. Retrieval Methodology

We conducted our search using the folloging keywords:

- **mental health and neurological disorders:** mental health, depression, mental health issues, alzheimer, anxiety, parkinson's, bipolar disorder, stress, dementia, somatic symptoms disorder;

- **speech dataset:** datasets, speech, corpus

As relevant scientific venues, we surveyed the following journals and conference proceedings.

- International Speech Communication Association Archive

- IEEE International Conference on Acoustics, Speech, and Signal Processing

- Language Resources and Evaluation Conference

- ACL Workshop on Computational Linguistics and Clinical Psychology

- International Conference on Machine Learning and Applications

- Clinical NLP Workshop

- ACM on Interactive, Mobile, Wearable and Ubiquitous Technologies

- ACM Multimedia

- IEEE Journal of Biomedical and Health Informatics

- IEEE Transactions on Affective Computing

- IEEE Engineering in Medicine and Biology Society

- American Journal of Speech-Language Pathology

- Frontiers in Neuroscience

- General Psychiatry Journal